\documentclass[10pt,twocolumn,letterpaper]{article}

\usepackage{iccv}
\usepackage{times}
\usepackage{epsfig}
\usepackage{graphicx}
\usepackage{amsmath}
\usepackage{amssymb}

\usepackage[accsupp]{axessibility}
\usepackage{microtype}
\usepackage{subfigure}
\usepackage{booktabs} 
\usepackage{soul}

\usepackage[ruled,vlined]{algorithm2e}

\usepackage{xspace}
\usepackage{multirow}
\usepackage{caption}
\usepackage{sidecap}
\usepackage{array}

\newcommand{\ours}{LGCL\xspace}
\newcommand{\method}{DualPrompt\xspace}

\newcommand{\myparagraph}[1]{\vspace{2pt}\noindent{\bf #1}}
\def\pp{\mathbf{P}}

\usepackage[pagebackref=true,breaklinks=true,colorlinks,bookmarks=false]{hyperref}

\iccvfinalcopy 

\def\iccvPaperID{3626} 

\ificcvfinal\fi

\begin{document}

\title{Introducing Language Guidance in Prompt-based Continual Learning}

\author{Muhammad Gul Zain Ali Khan$^{1,2}$ \quad Muhammad Ferjad Naeem$^{3}$\quad Luc Van Gool$^{3}$\quad Didier Stricker$^{1,2}$\\ Federico Tombari$^{4,5}$\quad Muhammad Zeshan Afzal$^{1,2}$\\
{\small $^1$RPTU\quad $^2$DFKI\quad $^3$ETH Zürich\quad $^4$TUM\quad $^5$Google}
}

\maketitle
\ificcvfinal\thispagestyle{empty}\fi

\begin{abstract}
   Continual Learning aims to learn a single model on a sequence of tasks without having access to data from previous tasks. The biggest challenge in the domain still remains catastrophic forgetting: a loss in performance on seen classes of earlier tasks. Some existing methods rely on an expensive replay buffer to store a chunk of data from previous tasks. This, while promising, becomes expensive when the number of tasks becomes large or data can not be stored for privacy reasons. As an alternative, prompt-based methods have been proposed that store the task information in a learnable prompt pool. This prompt pool instructs a frozen image encoder on how to solve each task. While the model faces a disjoint set of classes in each task in this setting, we argue that these classes can be encoded to the same embedding space of a pre-trained language encoder. In this work, we propose Language Guidance for Prompt-based Continual Learning~(\ours) as a plug-in for prompt-based methods. \ours is model agnostic and introduces language guidance at the task level in the prompt pool and at the class level on the output feature of the vision encoder. We show with extensive experimentation that \ours consistently improves the performance of prompt-based continual learning methods to set a new state-of-the art. \ours achieves these performance improvements without needing any additional learnable parameters.
\end{abstract}

\begin{figure}[t]
    \centering
    \includegraphics[width= \columnwidth]{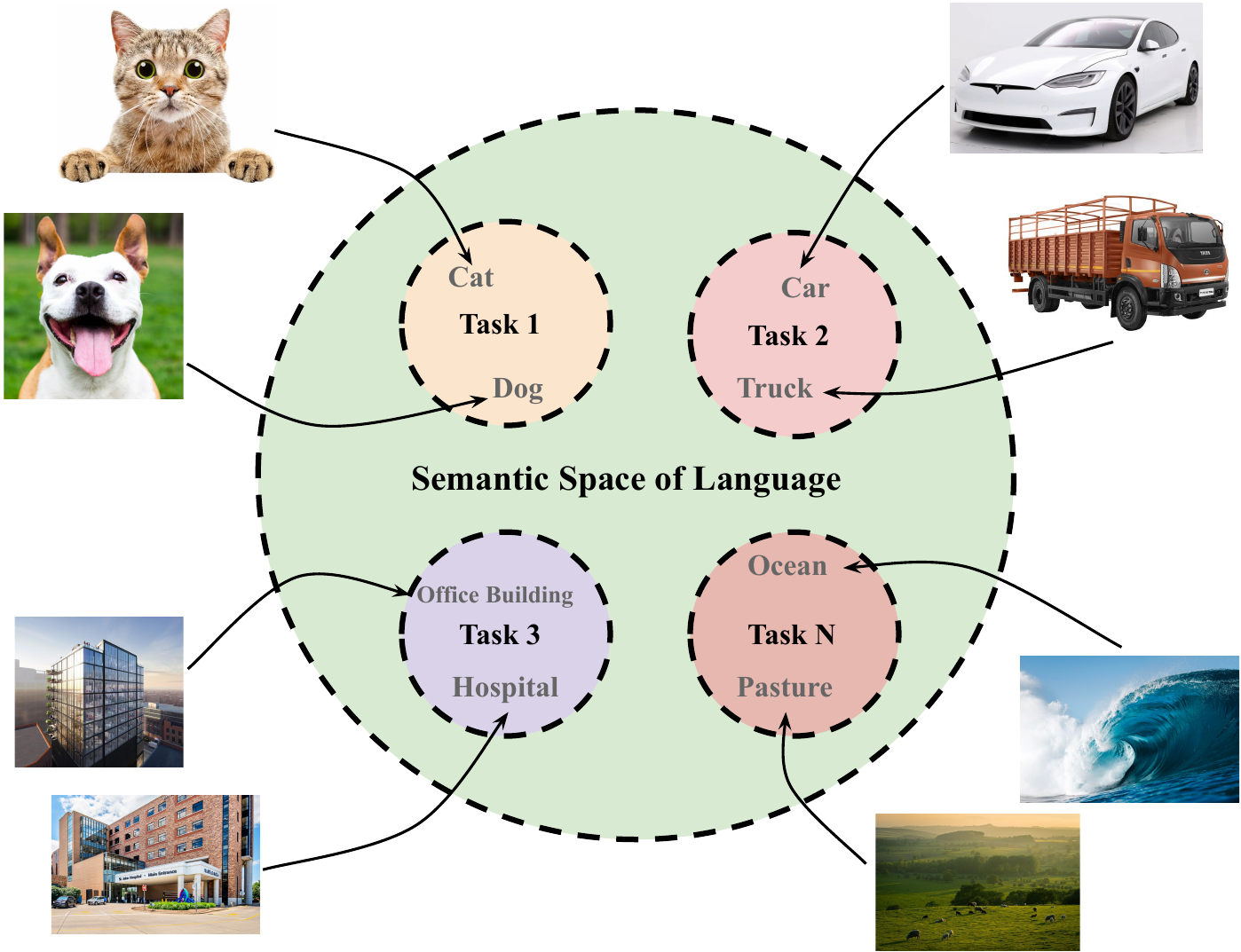}
    
    \vspace{-5pt}
    \label{fig:teaser}
    \caption{
    In Computer Vision, Continual Learning in the class incremental setting aims to learn a single model on a sequence of tasks where each task consists of disjoint classes. While each task represents a disjoint set of classes, we argue that they can be mapped to the same semantic space of a pretrained language encoder. Based on this principle, we propose to introduce language guidance in a continual learner to mitigate catastrophic forgetting.
    }
\end{figure}
\section{Introduction}
In Class Incremental Continual Learning, we task a model to learn a sequence of non-overlapping tasks consisting of new classes being introduced at each task. This presents a challenge different from the common supervised learning setting as the data distribution is continuously changing, and the independent and identically distributed (i.i.d.) data assumption does not hold. As a result, a model trained with our usual training recipe of optimising a loss function on incoming data leads to catastrophic forgetting~\cite{mccloskey1989catastrophic} \ie, the model forgets the previously seen classes since the loss only incentivises performance on the current task.
There have been several attempts to address this challenge. One popular line of works aims to identify model parameters most important for performance on each task and prevent them from changing too much through subsequent tasks~\cite{kirkpatrick2017overcoming, zenke2017continual, li2017learning, aljundi2018memory}. These regularization-based methods, however, achieve sub-optimal performance as we move to very complex tasks where the model needs to share parameters between different tasks to learn a robust representation.

Another line of work takes a very simple but effective approach of storing a chunk of training data. Rehearsal-based methods~\cite{chaudhry2018efficient, chaudhry2019tiny, hayes2019memory} maintain a rehearsal buffer which is a finite set of training data stored across each task. The key intuition is that to prevent forgetting on previously seen classes, the model simply uses the examples in the rehearsal buffer when optimising for new tasks. However, these methods require large buffer sizes as the number of tasks increases and hence become expensive. Moreover, they have been criticised for being impractical in real-world settings where privacy concerns prevent storing data. Architecture-based methods~\cite{rusu2016progressive, yoon2017lifelong, li2019learn, loo2020generalized, rao2019continual, zhao2022deep} take an orthogonal approach where these works reserve specialised parts of the network for each task and take an approach similar to multi-task learning. However, this can bring a significant increase in the number of learnable parameters. Moreover, this requires knowing the task identity at test time to select the relevant network module for each task which is not a realistic setting.

Recently, Learning to Prompt~(L2P)\cite{Wang2021l2p} has proposed an exciting new direction for continual learning. Instead of learning the parameters of the model for each new task, the authors propose to use a pre-trained vision encoder and learn the prompts that can instruct this pre-trained model to solve new tasks. This technique is called Prompt Tuning and is popularized by its success in Natural Language Processing~(NLP). A learned prompt instructs the model on how to solve a new task using the wide set of knowledge it has stored during pre-training. These methods have shown incredible performance boosts at a fraction of learnable parameters. L2P initialises a learnable prompt pool where each prompt is attached with a learnable key. The authors propose to use the $\mathtt{CLS}$ feature from the pre-trained vision encoder to perform a lookup with this learnable pool. The selected prompts are then appended with the patchwise embeddings of the image into the pre-trained model, and the output representations of the selected prompt tokens are used to learn a linear classification layer. This surprisingly simple formulation has brought impressive performance gains without the need to store any data in a rehearsal buffer. 

While the sequence of data from each task in continual learning has a changing distribution, we argue that the classes of each task can be mapped to the same semantic space. In this work, we argue that language represents such a robust representation space where all tasks can be sufficiently mapped to. Hence if we encode the features of the continual learner to map to a semantic space of language, this can present an avenue to mitigate catastrophic forgetting and result in a more robust continual learner. We use this insight to develop a novel method \textbf{Language Guidance for Prompt-based Continual Learning~(\ours)} that can introduce language guidance into any prompt-based continual learning method. We achieve this by introducing language guidance at two levels. First, we introduce the task-level language guidance by incentivising the model to map the learnable keys of the prompt pool into a shared language representation of all classes in the task. Secondly, we incentivise the model to map the output features of the visual encoder after prompting it to align with the language representation of its respective class. The model learns a robust representation of all tasks by aligning these representations with a pre-trained semantic space of language. 

Our contributions are as follows:
1) We present a novel perspective that entails introducing language guidance in continual learning to mitigate catastrophic forgetting.
2) We propose Language Guidance for Prompt-based Continual Learning~(\ours), a novel method that introduces language guidance in prompt-based continual learning methods.
3) Without any additional learnable parameters or extra memory requirements at inference, \ours improves the performance of prompt-based continual learning methods and achieves state-of-the-art performance on two challenging continual learning benchmarks.



\section{Related Work}
\myparagraph{Continual Learning} tasks a model to learn a sequence of tasks while mitigating catastrophic forgetting~\cite{mccloskey1989catastrophic}. Methods in continual learning have traditionally been divided into three categories, namely regularization-based methods, rehearsal-based methods, and architecture-based methods.
Regularization-based methods~\cite{kirkpatrick2017overcoming, zenke2017continual, li2017learning, aljundi2018memory} aim to find important parameters for each task and limit their plasticity in future tasks by adjusting the learning rate. These methods work without storing any labelled examples; however, they are unable to achieve satisfactory performance in challenging and complex datasets~\cite{mai2021online,rebuffi2017icarl, wu2019large}.

Rehearsal-based methods~\cite{chaudhry2018efficient, chaudhry2019tiny, hayes2019memory} maintain a buffer to save data from older tasks and use it for training while future tasks become available. Several works improve upon it with training tricks like knowledge distillation~\cite{rebuffi2017icarl, wu2019large, chaudhry2020using, buzzega2020dark} and self-supervised learning~\cite{cha2021co2l, pham2021dualnet}. Rehearsal-based methods address catastrophic forgetting by simply retraining on stored data from all tasks the model has seen at any given stage. Although conceptually very simple, these methods have been very competitive and consistently rank among state-of-the-art~\cite {parisi2019continual, mai2021online}. However, these methods suffer from performance degradation as the replay buffer gets smaller or the number of classes increases significantly. Moreover, these methods can not be used when data privacy is a concern~\cite{shokri2015privacy}.

Architecture-based methods aim to specialize parts of the model for each task. These modules are added as additional blocks~\cite{rusu2016progressive, yoon2017lifelong, li2019learn, loo2020generalized, rao2019continual, zhao2022deep}, or specialising task specific sub-networks~\cite{mallya2018packnet, serra2018overcoming, wang2020learn, ke2020continual}. Since these models specialise parts of the model for each task, they often require the task identity as an input to the model at test time which limits their use in realistic class-incremental and task-agnostic settings. Some methods infer task identity from the data~\cite{wortsman2020supermasks}, while others infer it using a rehearsal buffer~\cite{yan2021dynamically, pham2021dualnet}. However, these methods require significantly more learnable parameters, often as many as the core model. 
Prompt-based methods~\cite{Wang2021l2p, wang2022dualprompt} have recently emerged as a new exciting fourth direction in continual learning. These methods use a pre-trained feature extractor and learn each task as a set of prompts that specialise the pre-trained model for the task. These methods are highly parameter efficient as prompts are small sequences of learnable tokens. These methods achieve this by encoding the task information in the learnable prompts rather than storing input data. Moreover, these methods do not require the task identity as input, thanks to a clever lookup formulation conditioned on the input to select the prompt.

\myparagraph{Prompt Learning} has emerged as a popular transfer learning technique in Natural Language Processing~(NLP). Instead of retraining the model, prompt learning learns a set of prompts that instructs the pre-trained model to process the new task. To this end, several works introduce prompts as learnable tokens achieving impressive performance on transfer learning~\cite{lester2021power,li2021prefix}. These methods are incredibly efficient with respect to learnable parameters compared to competitors~\cite{wang2020k,pfeiffer2020adapterfusion,hu2021lora}.

\myparagraph{Language Guidance} has been extensively explored in various vision tasks, including open set learning~\cite{Radford2021clip, ghiasi2022scaling, wang2022s}, zero-shot learning~\cite{naeem2022i2dformer, naeem2022i2mvformer, cge, cape}, and metric learning~\cite{roth2022integrating}. Methods in open set learning~\cite{Radford2021clip, ghiasi2022scaling} learn a vision encoder that can map to the same embedding space as language. The model can then generalise to new classes by generating the embeddings of the class names without requiring labelled visual data. Methods in zero-shot learning use word embeddings from pre-trained language models~\cite{yamada2018wikipedia2vec,GloVe,socher2013zero} and knowledge graphs~\cite{wang2018zero,kampffmeyer2019rethinking,bucher2017generating, cge, cocge} to encode semantic similarities between seen and unseen classes. Unseen classes can then be inferred by measuring a distance metric between a vision encoding and a language feature from a pre-trained model. Integrating language supervision in vision models allows the model to adapt to new classes efficiently, as these classes lie in the same semantic space as previously seen classes.

Our method lies at the intersection of prompt-based continual learning and language guidance. To the best of our knowledge, we provide the first method for integrating language guidance in prompt-based continual learning methods for challenging class-incremental continual learning.
\section{Background}

\begin{figure*}
    \centering
    \includegraphics[width=\textwidth]{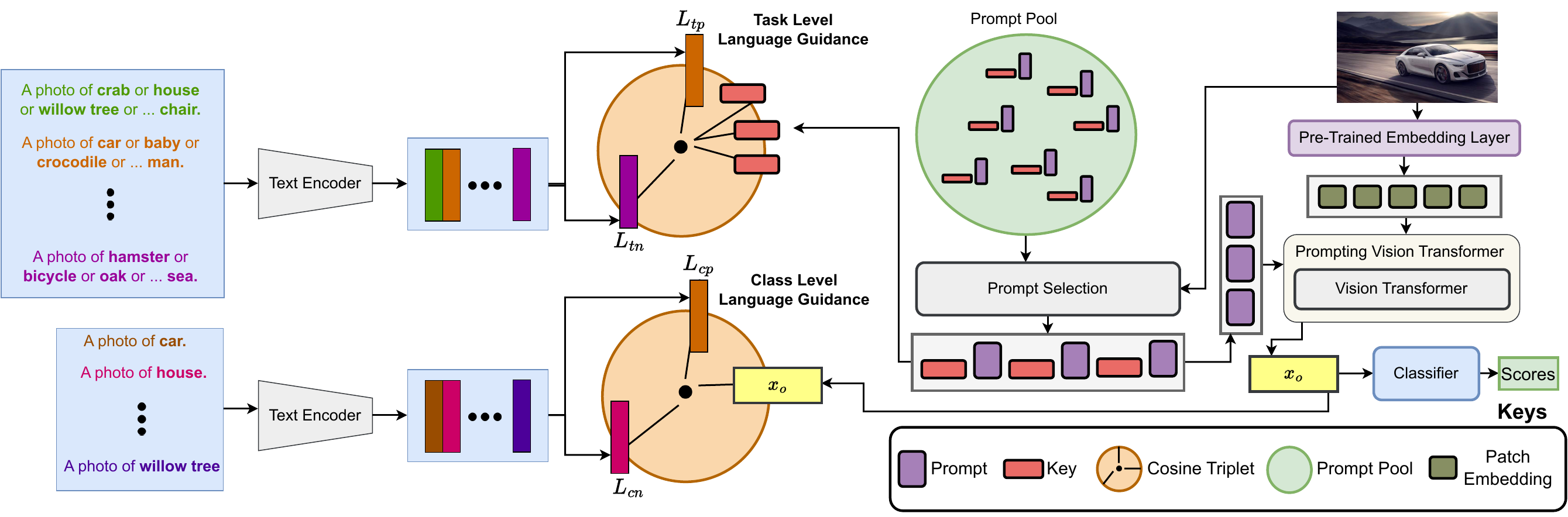}
    \caption{ 
    Our novel \textbf{Language Guidance for Prompt-based Continual Learning~(LGCL)} introduces Language Guidance in prompt-based continual learning methods. We introduce language guidance at two levels. At task level, we encode the language feature corresponding to the classes the model will encounter in the selected keys from the prompt pool. At class level, we encode the language feature corresponding the the ground truth class in the output feature of the vision transformer. Together the two modules improve the baseline prompt-based continual learning method and bring performance improvements without introducing additional learnable parameters.
    }
    \label{fig:model}
\end{figure*}


\myparagraph{Notations.}

\label{sec:notations}
Continual Learning aims to train a Machine Learning model on a stream of data from a sequence of tasks. We denote the sequence of tasks as $\mathcal{D}\ =\ \{\mathcal{D}_{1},\mathcal{D}_{2},...,\mathcal{D}_{\text{T}}\}$ where each task consists of tuples of input data $(x_{i}^{t},y_{i}^{t})$ where $x_i^t \in \mathcal{X}$ represents the input image in an RGB color space and $y_i^t \in \mathcal{Y}$ represents the corresponding label for the task $t$. In line with previous works, the tasks are non-overlapping, \ie images and labels are not repeated in subsequent tasks, and the model does not have access to the training data from the previous tasks. We focus on class incremental setting in which task identity is unknown at test time.
Moreover, we assume that a pre-trained feature extractor $\mathcal{F}$ is available for images and is kept frozen throughout the training~\cite{wang2022dualprompt,Wang2021l2p}. Similarly, we assume that a pre-trained feature extractor is available for language and is kept frozen throughout the training~\cite{Radford2021clip}. To be consistent with previous works, vision and language feature encoders are each independently trained. We consider the class incremental setting in which task boundaries are defined clearly, and task identity is unknown during test time~\cite{Pham2021DualNetCL}.


\subsection{Prompt-based Continual Learning}
Since our method aims to improve prompt-based continual learning methods, we provide an overview in this section. The vast majority of continual learning works maintain a Replay Buffer consisting of labelled samples of previous tasks. This buffer is used to avoid catastrophic forgetting by continuously training on previous tasks. However, rehearsal buffers are expensive to store and do not scale well to large dataset or a large number of tasks. Recently, prompt tuning has emerged as an alternative to rehearsal buffers. Methods in this direction~\cite{wang2022dualprompt,Wang2021l2p} use a pre-trained vision encoder and rely on prompt learning to learn the tasks continually, instead of replaying samples from previous tasks. This is achieved by storing the knowledge of each task in a learnable pool of prompts without explicitly defining a pool for each task.

Given a pre-trained image feature extractor $\mathcal{F}$, an image transformer, these methods aim to learn prompts that can be used to instruct the pre-trained model to solve the encoded task. Given an image $x$, they do a forward pass to extract the $\mathtt{CLS}$ token corresponding to the global feature of the image. This feature is used to look up the relevant prompt from the prompt pool, which we introduce in the next section.

\myparagraph{Introducing a learnable Prompt Pool.}
Prompt learning has emerged as a powerful technique in NLP to use a general pre-trained language model and re-purpose it for a downstream task by introducing a set of learnable tokens without changing the parameters of the pre-trained model. Prompt Tuning~\cite{lester2021power} introduces a set of learnable tokens for a pre-trained language model like T5~\cite{raffel2020exploring} to condition the pre-trained model to solve a new task. These tokens encode the task instructions and instruct the pre-trained model to solve the NLP task at hand~\cite{liu2023pre}. On the other hand, another form of utilizing learnable prompts is prefix tuning. In prefix tuning, the learnable prompt is appended to the keys and values of attention blocks~\cite{wang2022dualprompt}.

Introducing Prompting for Continual Learning involves some clever design choices. We want to utilize the prompt to fine-tune the internal representations of the vision transformer for our task-specific distribution without tuning the model parameters. The simplest approach is to learn one set of prompt tokens for each task capturing the task-specific information in its tokens. However, this has a significant limitation in that the model needs the task id as an input to select the correct prompt. Moreover, this does not allow the model to create a joint representation that can share similar information between tasks. Learning to Prompt~(L2P)~\cite{Wang2021l2p} instead cleverly introduces a pool of prompts where each prompt can encode knowledge without explicitly attaching it to a task. The prompt pool is defined as:
\begin{equation}
\label{eq:prompt_pool}
    \pp=\{P_1, P_2, \cdots, P_M\}, \quad \text{$M = $ total $\#$ of prompts},
\end{equation}
where $P_j \in \mathbb{R}^{L_p\times E}$ is a single prompt with token length $L_p$ and the same embedding size $E$ as $x_p$. Each prompt is attached to a learnable key $k_j$. Given the $\mathtt{CLS}$ feature corresponding to an input image $x$ as a query, the model can look up the relevant prompt encoding the knowledge for its task by a key query look up. Learning to Prompt~\cite{Wang2021l2p} uses top N prompts corresponding to the lookup as the selected prompts for tuning. The selected prompts are used as additional input to the pre-trained Vision Transformer, along with patch embeddings. Dual Prompt\cite{wang2022dualprompt} instead uses prefix-tuning and directly injects these prompts in the Multi-headed attention layers of the Vision Transformer by prepending the learnable prompt with keys and values of the Multi-Head Attention layer. We name the pre-trained frozen vision transformer prompted with learnable prompts the Prompting Vision Transformer. 

\myparagraph{Encoding task information in the selected prompts.} 
We define $x_o$ as the output feature of the Vision Transformer to be used for classification. In Dual Prompt\cite{wang2022dualprompt} this corresponds to the $\mathtt{CLS}$ feature of the Vision transformer after injecting the selected prompts. In L2P\cite{Wang2021l2p} this refers to the average pooled output of the tokens corresponding to the selected prompts. This feature is trained for classification with a supervised loss like Cross-Entropy for classification tasks. The training loss incentivises the model to store task-specific features in the prompt pool.
Moreover, since the prompt selection is dependent on the input image and not the task, the task representations are shared in the pool and evolve over the training period to encode the different tasks. Dual Prompt~\cite{wang2022dualprompt} improves upon L2P~\cite{Wang2021l2p} by additionally introducing a global learnable prompt which learns a shared representation across all tasks. This global prompt is used with the prompts as input to the Vision Transformer. 

\section{Language Guidance for Prompt-based Continual Learning~(LGCL).}

Continual Learning addresses the task of learning a changing distribution of data coming from different tasks. While the visual data of these tasks changes, their task definition or classification targets can lie in the same space of language. Language consists of a compact representation of the world and storing language cues like class names is available to a model for free as it has access to them from the current and previous tasks. We propose to integrate this language guidance into the prompt-based continual learning methods to further mitigate catastrophic forgetting. More specifically, we propose to use a text encoder $\mathcal{T}$ from a pre-trained model to encode the task knowledge and class knowledge into the prompt pool and learned features of the continual learner. \ours is a generic framework that can be incorporated into any prompt-based method for continual learning without requiring any additional learnable parameters. We give an overview of our method in Figure~\ref{fig:model}.

\subsection{Introducing Task Level Language Guidance}
Given the $t-th$ task, we denote the class names of classes represented in this task with the set $\mathcal{Y}^t$. The task $t$ involves correctly classifying the classes contained in $\mathcal{Y}^t$. Therefore, we propose to represent the language representation of the task as a prompt of class names as follows.
\begin{center}    
 ``A photo of \{\textbf{class 1}\} or \{\textbf{class 2}\} $\cdots$ or \{\textbf{class n}\}"\\
\end{center}
where \{\textbf{class 1}\},$\cdots$, \{\textbf{class n}\} are replaced with the class names of the task. The prompt is input to the pre-trained text encoder to extract the feature corresponding to the output token to represent $L_t \in \mathbb{R}^E$, the language representation of the task $t$ with embedding dimension $E$. Since prompt-based continual learning methods learn the lookup operation for selecting the prompts against learnable keys, we aim to encode the task definition in these keys. Given $P_s = \{P_{s_1}; \cdots; P_{s_N}\}$ are the $N$ prompts selected for the task $t$, with learnable keys $K_s = \{k_{s_1}; \cdots;  k_{s_N} \}$, we aim to encode the $L_t$ in these learnable keys. 
For each key $k \in \mathbb{R}^{E}$ in $K_s$, we compute the cosine similarity between the key and the language encoded task feature $L_t$ as follows:
\begin{equation}
    S(k, L_t) = \frac{k\cdot L_t}{|k||L_t|}
\end{equation}
We optimise the cosine similarity with a triplet loss to incentivise the model to align the selected keys close to the language representation of their respective task and away from the language representation of other tasks. Given the task $t$, the model only has access to the task definitions of the current task and the tasks before $t$. Therefore, when optimising the loss for the current task $t$, $L_{tp}$ denotes the language feature of the task $t$ as the positive and the language features of the previous tasks are randomly sampled as the negative $L_{tn}$ in each optimisation step. For a selected key $k$, we optimise the following loss:
\begin{equation}
    \mathcal{L}_{task}(k, L_{tp}, L_{tn})=1-S(k,L_{tp})+S(k,L_{tn})
    \label{eq:cosine_triplet}
\end{equation}

By aligning the lookup keys with the language feature of the task, the model learns a feature representation of keys that comes from the same distribution of language and is less likely to diverge between tasks while training. Since the performance of prompt-based continual learning methods depends on the correctness of the selected prompts, learning better keys can allow for better performance.

\subsection{Introducing Class Level Language Supervision.}
The prompt pool represents the task-level knowledge for the model. We further want to guide the class-level feature of the image with language. For a given training sample $(x, y)$ consisting of image $x$ with label $y$, we take the class name for $y$ and represent it in language as the following prompt:
\begin{center}
``A photo of \{\textbf{class name}\}"\\
\end{center}
Similar to the last module, the prompt is input to the pre-trained text encoder to extract the feature corresponding to the output token to represent $L_c \in \mathbb{R}^E$, the language representation of class $y$ with embedding dimension $E$. We want to encode this language representation in the output feature $x_o \in \mathbb{R}^E$ of the vision transformer used for classification. This feature aims to represent the class-level information of the task. We introduce language guidance in this feature representation through a cosine triplet loss similar to the last module. Our positive example consists of the language-encoded feature of class $y$ as $L_{cp}$. For the negative example, we randomly sample a class from the classes of the previous tasks as $L_{cn}$ for each optimization step. We optimise the following loss for introducing language guidance in our continual learner:
\begin{equation}
    \mathcal{L}_{class}(x_o, L_{cp}, L_{cn})=1-S(x_o,L_{cp})+S(x_o,L_{cn})
    \label{eq:cosine_tripletclass}
\end{equation}
By aligning output image features to the classwise language representation, we incentivise the model to map to the same semantic space of the pre-trained language encoder across each task. We keep all other aspects of the baseline methods the same from their respective authors. 

\myparagraph{Inference.}
The model does not require language guidance at inference, and the baseline prompt-based methods can be used with their original formulation. The $x_o$ features are extracted and classified with a linear layer.

\begin{table*}[t!]
\begin{center}
\scalebox{0.9}{
\begin{tabular}{l|c|cc|c|cc}
\toprule 
 \multirow{1}{*}{\textbf{Method}} & \multirow{1}{*}{\textbf{Buffer size}} & \multicolumn{2}{c|}{\textbf{Split CIFAR-100}} & \multirow{1}{*}{\textbf{Buffer size}} & \multicolumn{2}{c}{\textbf{Split ImageNet-R}} \\
& &  Avg. Acc ($\uparrow$) & Forgetting ($\downarrow$) & & Avg. Acc ($\uparrow$) & Forgetting ($\downarrow$) \\
\midrule
   ER \cite{chaudhry2019tiny} & \multirow{6}{*}{1000} & 67.87\scriptsize{$\pm$0.57} & 33.33\scriptsize{$\pm$1.28} & \multirow{6}{*}{1000} & {55.13\scriptsize{$\pm$1.29}} & 35.38\scriptsize{$\pm$0.52} \\
   BiC \cite{wu2019large} & & 66.11\scriptsize{$\pm$1.76} & 35.24\scriptsize{$\pm$1.64} && 52.14\scriptsize{$\pm$1.08} & 36.70\scriptsize{$\pm$1.05} \\
  GDumb \cite{prabhu2020gdumb} & & 67.14\scriptsize{$\pm$0.37} & - && 38.32\scriptsize{$\pm$0.55} & - \\
 DER++ \cite{buzzega2020dark} & & 61.06\scriptsize{$\pm$0.87} & 39.87\scriptsize{$\pm$0.99} && 55.47\scriptsize{$\pm$1.31} & 34.64\scriptsize{$\pm$1.50} \\
 Co$^2$L \cite{cha2021co2l} & & 72.15\scriptsize{$\pm$1.32} & 28.55\scriptsize{$\pm$1.56} && 53.45\scriptsize{$\pm$1.55} & 37.30\scriptsize{$\pm$1.81} \\
 \midrule
 ER \cite{chaudhry2019tiny} &\multirow{6}{*}{5000 }& 82.53\scriptsize{$\pm$0.17} & 16.46\scriptsize{$\pm$0.25} &\multirow{6}{*}{5000 }& 65.18\scriptsize{$\pm$0.40} & 23.31\scriptsize{$\pm$0.89} \\
 BiC \cite{wu2019large} & & 81.42\scriptsize{$\pm$0.85} & 17.31\scriptsize{$\pm$1.02} && 64.63\scriptsize{$\pm$1.27} & 22.25\scriptsize{$\pm$1.73} \\
 GDumb \cite{prabhu2020gdumb} & & 81.67\scriptsize{$\pm$0.02} & - && 65.90\scriptsize{$\pm$0.28} & -  \\
 DER++ \cite{buzzega2020dark} & & 83.94\scriptsize{$\pm$0.34} & 14.55\scriptsize{$\pm$0.73} && 66.73\scriptsize{$\pm$0.87} & 20.67\scriptsize{$\pm$1.24} \\
 Co$^2$L \cite{cha2021co2l} & & 82.49\scriptsize{$\pm$0.89} & 17.48\scriptsize{$\pm$1.80} && 65.90\scriptsize{$\pm$0.14} & 23.36\scriptsize{$\pm$0.71} \\
 \midrule
 FT-seq &  & 33.61\scriptsize{$\pm$0.85} & 86.87\scriptsize{$\pm$0.20} & \multirow{5}{*}{0} & 28.87\scriptsize{$\pm$1.36} & 63.80\scriptsize{$\pm$1.50}\\
 EWC \cite{kirkpatrick2017overcoming} & & 47.01\scriptsize{$\pm$0.29} & 33.27\scriptsize{$\pm$1.17} && 35.00\scriptsize{$\pm$0.43} & 56.16\scriptsize{$\pm$0.88} \\
 LwF \cite{li2017learning} & 0 & 60.69\scriptsize{$\pm$0.63} & 27.77\scriptsize{$\pm$2.17} && 38.54\scriptsize{$\pm$1.23} & 52.37\scriptsize{$\pm$0.64} \\
 \midrule
 {L2P} \cite{wang2021learning} & & {83.86\scriptsize{$\pm$0.28}} & {7.35\scriptsize{$\pm$0.38}} & &{61.57\scriptsize{$\pm$0.66}} & {9.73\scriptsize{$\pm$0.47}} \\
 \bf{L2P + \ours} (Ours) & 0 & \underline{{84.33}\scriptsize{$\pm$0.06}} & \underline{5.83\scriptsize{$\pm$0.23}} & 0 &\underline{62.51\scriptsize{$\pm$0.05}} & \underline{8.9\scriptsize{$\pm$0.17}} \\
 \midrule
{\method} &  &  86.51\scriptsize{$\pm$0.33} &  5.16\scriptsize{$\pm$0.09} &  &  68.13\scriptsize{$\pm$0.49} &  4.68\scriptsize{$\pm$0.20}  \\
  \bf{\method + \ours (Ours)} & 0& \underline{\bf 87.23\scriptsize{$\pm$0.21}} & \underline{\bf 5.10\scriptsize{$\pm$0.15}} & 0 & \underline{\bf 69.46\scriptsize{$\pm$0.04}} & \underline{\bf 4.2\scriptsize{$\pm$0.06}}  \\
\midrule
Upper-bound & -& 90.85\scriptsize{$\pm$0.12} & - & - & 79.13\scriptsize{$\pm$0.18} & - \\
\bottomrule
\end{tabular}
}
\caption{\textbf{Results on class incremental learning.} We compare \ours with baseline and previous methods. Following \cite{wang2022dualprompt}, we group methods by buffer size. Our method is proposed for prompt-based methods like ~\cite{wang2022dualprompt,Wang2021l2p} and therefore, require no rehearsal buffer. We observe \ours outperforms previous baseline methods in Split-ImageNet-R~\cite{wang2022dualprompt} and Split CIFAR-100~\cite{Krizhevsky2009cifar100} consistently.}
\label{tab:mainresults}
\end{center}
\vspace{-.6cm}
\end{table*} 

\section{Experiments}
\myparagraph{Experiment Protocol.} 
Consistent with previous works\cite{Wang2021l2p, pham2021dualnet}, we used ViT B/16\cite{vit} pre-trained on ImageNet 1k as our Image feature extractor. This is kept frozen during training. On the language side, we use the text transformer of CLIP L/14\cite{Radford2021clip} for our main experiments. We use the Adam optimizer ~\cite{kingma2014adam} with $\beta_1 = 0.9$ and $\beta_2 = 0.999$. We set the batch size to 24 for Dual Prompt~\cite{wang2022dualprompt} and 16 for L2P~\cite{Wang2021l2p}. 
We train on one A100-40GB GPU with the code released by the authors of each method. Input images are resized to $224 \times 224$ and normalized to the range of [0,1].  We follow ~\cite{Buzzega2020DER,Wang2021l2p,wang2022dualprompt} and train for multiple epochs. For L2P~\cite{Wang2021l2p} Split CIFAR-100~\cite{Krizhevsky2009cifar100}, we train 5 epochs, for L2P~\cite{Wang2021l2p} Split ImageNet-R~\cite{wang2022dualprompt} we train 50 epochs. We train Dual Prompt~\cite{wang2022dualprompt} for 20 epochs on Split CIFAR-100~\cite{Krizhevsky2009cifar100} and for 50 epochs on Split ImageNet-R~\cite{wang2022dualprompt}. 
For comparison with state-of-the-art, we use the widely adopted Average accuracy~(higher is better) and Forgetting~(lower is better) to compare model performance~\cite{lopez2017gradient,chaudhry2018efficient,mai2021online}.
Since prompt-based continual learning is a very recent development, we use the two most recent baselines Learning to Prompt\cite{Wang2021l2p} and Dual Prompt\cite{wang2022dualprompt} and incorporate \ours in training. 
To make the comparison fair, we use the same hyperparameters for L2P~\cite{Wang2021l2p} and Dual Prompt~\cite{wang2022dualprompt} as provided in their code repositories and paper. We do not perform any hyperparameter optimisation for \ours. Since our $\mathcal{L}_{task}$ and $\mathcal{L}_{class}$ require negatives from previous tasks, they are used once the first task is learned.
We compare with regularization and rehearsal-based methods in Table~\ref{tab:mainresults} as these can be trained with the same transformer-based visual encoder. We further compare with architecture-based methods in Table~\ref{tab:diffresults}. Since these models are trained with different visual encoders, we compare performance against them as a difference from supervised performance.

\subsection{Datasets}
\myparagraph{Split Imagenet-R.} The split ImageNet-R~\cite{wang2022dualprompt} is built on ImageNet-R~\cite{Hendrycks2020split-imagenet-r}. It contains 200 classes that are split into 10 disjoint tasks, with each task containing 20 classes. The dataset is divided into 24,000 training images and 6000 test images. Split ImageNet-R~\cite{wang2022dualprompt} has more diversity in the images and is closer to the complicated real-world images. 

\myparagraph{Split CIFAR-100.} Split CIFAR-100 is a widely used dataset for continual learning. Split CIFAR-100 is made of 10 disjoint tasks with 10 classes per task taken from the original CIFAR-100~\cite{Krizhevsky2009cifar100}. Compared to Split ImageNet-R, it is a simpler dataset for classification, however, it is sufficient to expose the large forgetting rate of CL methods in class-incremental learning~\cite{wang2022dualprompt}.

\subsection{Comparison with State-of-the Art}
We compare with various regularization-based, rehearsal based and prompt-based methods for continual learning in Table~\ref{tab:mainresults}. We observe that Dual Prompt\cite{wang2022dualprompt} paired with our model \ours achieves the best results and sets a new state-of-the-art. We further observe that prompt-based methods significantly outperform regularization-based and rehearsal-based methods on both datasets.

As we compare Dual Prompt with \ours + Dual Prompt\cite{wang2022dualprompt}, we see that the introduction of language guidance brings a decent improvement. On Split CIFAR-100, \ours improves Dual Prompt by 0.72\% on average accuracy, the measure of average performance across all tasks. Similarly, on Split ImageNet-R, \ours improves Dual Prompt by an impressive 1.33\%.


As we compare the second best method L2P\cite{Wang2021l2p}, we see that the introduction of \ours brings similar performance improvements. On Split CIFAR-100, the method sees an improvement of 0.47\% in average accuracy and an impressive 1.52\% on the forgetting metric. Similarly, on Split ImageNet-R, the method sees an improvement of 0.94\% in average accuracy and 0.83\% in forgetting. This validates our hypothesis that the introduction of language guidance with \ours improves model performance without requiring any additional learnable parameters. 

\begin{table*}[t!]
\small
\label{table:architecture}
\begin{minipage}{\textwidth}
\begin{center}
\scalebox{1.0}{
\begin{tabular}{l|c|lc|c|>{\centering\arraybackslash}p{1.8cm}>{\centering\arraybackslash}p{1.8cm}}
\toprule 
 \multirow{2}{*}{\textbf{Method}} & \multirow{2}{*}{\textbf{Backbone}} & \multirow{2}{*}{\textbf{Avg. Acc ($\uparrow$)}} & \multirow{2}{*}{\textbf{Diff ($\downarrow$)}} & \multirow{2}{*}{\textbf{Buffer size}} &  \multicolumn{2}{c}{\textbf{Additional Parameters}} \\
& &  & & & MB & \% \\
\midrule
  Upper-bound & \multirow{5}{*}{ResNet18}& 80.41$^\dagger$ & - & - & - & - \\
   SupSup~\cite{wortsman2020supermasks} & & 28.34\scriptsize{$\pm$2.45}$^\ddagger$ & 52.07 & 0 & 3.0 & 6.5\% \\
   DualNet~\cite{pham2021dualnet} & & 40.14\scriptsize{$\pm$1.64}$^\ddagger$ & 40.27 & 1000 & 5.04 & 10.9\% \\
  RPSNet~\cite{rajasegaran2019random} & & 68.60$^\dagger$ & 11.81 & 2000 & 181 & 404\% \\
 DynaER~\cite{yan2021dynamically} & & 74.64$^\dagger$ & 5.77 & 2000 & 19.8 & 43.8\% \\
 \midrule
 Upper-bound & \multirow{2}{*}{ResNet152}& 88.54$^\dagger$ & - & - & - & - \\
 DynaER~\cite{yan2021dynamically} & & 71.01\scriptsize{$\pm$0.58}$^\ddagger$ & 17.53 & 2000 & 159 & 68.5\% \\
 \midrule
Upper-bound & \multirow{5}{*}{ViT-B/16}& 90.85\scriptsize{$\pm$0.12}$^\ddagger$ & - & - & - & - \\
 {L2P} \cite{wang2021learning} &  & {83.86\scriptsize{$\pm$0.28}}$^\ddagger$ & 6.99 & 0 & 1.94 & 0.56\% \\
  {L2P + \ours} (Ours) &  & \underline{84.33\scriptsize{$\pm$0.06}}& \underline{6.52} & 0 & 1.94 & 0.56\% \\
{\method} & &  86.51\scriptsize{$\pm$0.33}$^\ddagger$ &  4.34 & 0 & \bf 1.90 &  \bf 0.55\%  \\
 \bf{\method + \ours} (Ours) & & \underline{\bf 87.23\scriptsize{$\pm$0.21}} & \underline{\bf 3.45} & 0 & \bf 1.90 &  \bf 0.55\%  \\
\bottomrule
\end{tabular}
}
\\
\scriptsize{$^\dagger$Reported by the original papers. $^\ddagger$ Reproduced using their original codebases.}
\end{center}
\end{minipage}
\caption{\textbf{Comparison with Architecture Based methods on Split-CIFAR-100.} The Upper-Bound denotes the model performance when trained in a fully supervised, non-continual setting, i.e., with access to all tasks at the same time. Following ~\cite{wang2022dualprompt}, we use \texttt{Diff = Upper-Bound Acc - Method Acc} (lower is better). This measures how close the model is to the supervised performance across different model backbones. We observe \ours outperforms baseline methods and consistently improves the performance of prompt-based continual learning methods.}
\label{tab:diffresults}
\end{table*}

We compare the performance of \ours and prompt-based methods with architecture-based methods in Table~\ref{tab:diffresults}. These methods are trained on different backbones. To be consistent with previous works\cite{wang2022dualprompt, Wang2021l2p}, we report the difference between supervised performance and the model performance as the metric.
We observe from the Table \ref{tab:diffresults} that \ours again sets a new state-of-the-art in this setting too. \ours consistently outperforms methods with big buffer sizes. As we compare the performance of Dual Prompt with and without \ours, we again notice an improvement. \ours further pushes Dual Prompt towards the upper bound supervised performance with a difference of only 3.45\% from the supervised performance. 
\begin{figure}
    \centering
    \begin{subfigure}{}
        \includegraphics[width=0.6\linewidth]{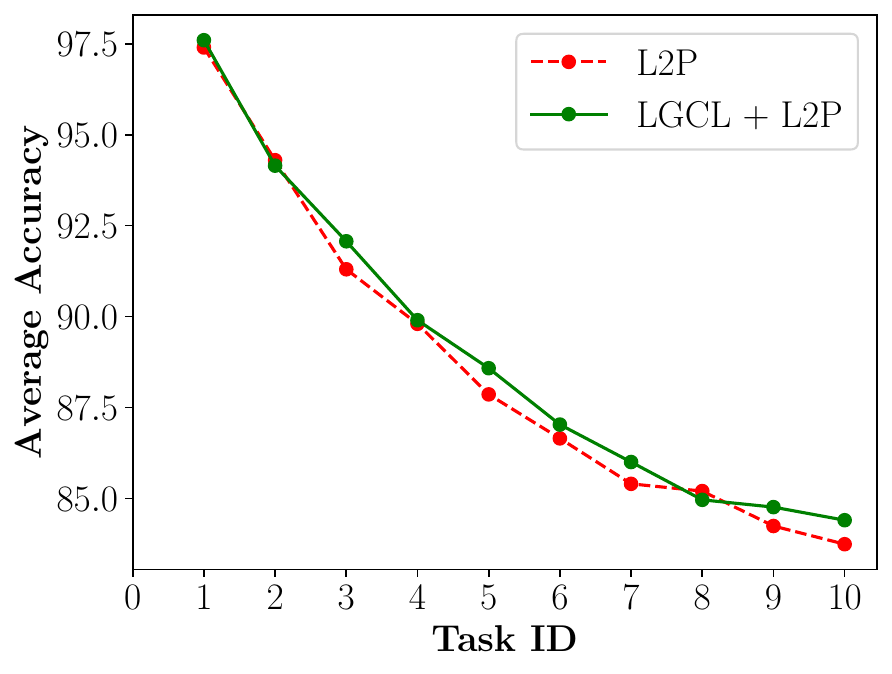}
        \vspace{-8pt}
        \caption{Comparison of average accuracy at each task of L2P~\cite{Wang2021l2p} + \ours. We observe that \ours on average prevents a drop in performance across tasks.}
        \label{fig:l2p_per_task}
    \end{subfigure}%
    \begin{subfigure}{}
        \includegraphics[width=0.6\linewidth]{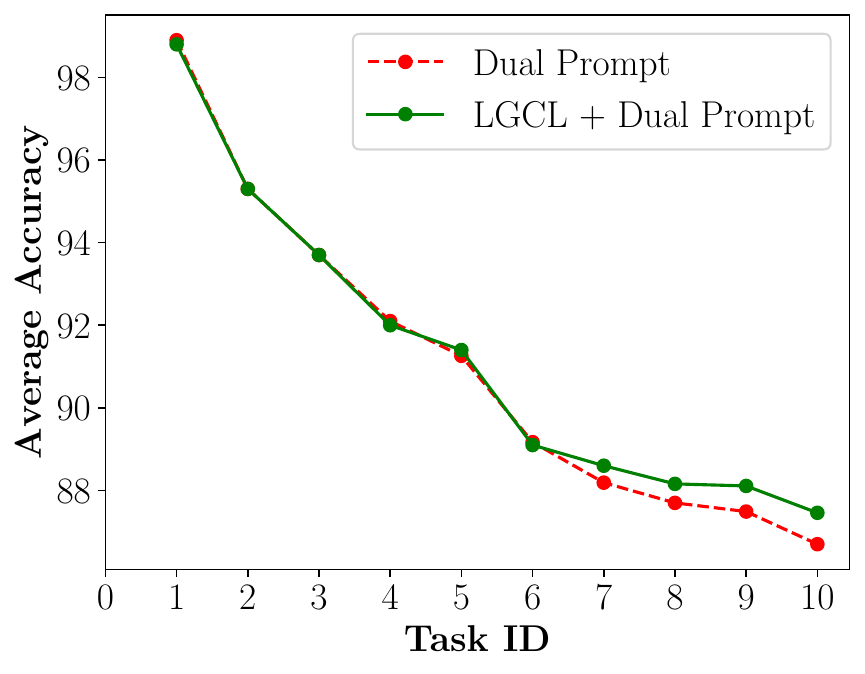}
        \vspace{-8pt}
        \caption{Comparison of average accuracy at each task of Dual Prompt ~\cite{wang2022dualprompt} + \ours. We observe that \ours on average prevents a drop in performance across tasks.}
        \vspace{-8pt}
        \label{fig:dualprompt_per_task}
    \end{subfigure}%
    \label{fig:graphs_per_task}
\end{figure}


\subsection{Ablation on the components of \ours.}
We test each component of our model \ours on the challenging Split Imagenet-R dataset and report the results in Table~\ref{tab:ablateloss} for both L2P~\cite{Wang2021l2p} and Dual Prompt~\cite{wang2022dualprompt}. Comparing rows a) and b), the introduction of $\mathcal{L}_{task}$ leads to a slight improvement in both L2P and Dual Prompt. Comparing rows a) and c), we see a similar conclusion where the introduction of class-level language loss leads to a decent improvement in both datasets. Comparing rows b) and c), we observe that class-level language guidance leads to a bigger improvement than only task-level language guidance. Finally, as we observe from row d), our full model \ours uses both task-level and class-level language supervision in training and achieves more than a full point improvement on both baseline methods indicating the effectiveness of both modules of our model. We, therefore, conclude that the introduction of both task-level and class-level language guidance is complementary and consistently improves the prompt-based continual learning methods. We once again want to emphasise that this improvement is achieved without introducing any additional learnable parameters.

\subsection{Per task performance improvement.}
We plot the Average Accuracy of the model through the ten tasks in Figure~\ref{fig:l2p_per_task} for L2P and Figure~\ref{fig:dualprompt_per_task} for Dual Prompt on Split CIFAR-100 dataset. As we compare the model performance with and without \ours on L2P in Figure~\ref{fig:l2p_per_task}, we see that the model with language guidance is slow in dropping accuracy as each additional task is introduced. We see that performance improvement of introducing language can be observed at each training stage. We similarly compare the performance of Dual Prompt with and without \ours in Figure~\ref{fig:dualprompt_per_task} and see a similar trend where the introduction of language guidance results in smaller drops in performance as the model is trained for more tasks. This again validates our hypothesis that the introduction of language guidance can mitigate catastrophic forgetting without including any additional trainable parameters.

{
\setlength{\tabcolsep}{4pt}
\renewcommand{\arraystretch}{1.2} 
\begin{table}[t]
\setlength{\aboverulesep}{0pt}\setlength{\belowrulesep}{0pt}
    \centering
    {
    \begin{tabular}{l cc cc cc}
    \toprule
    & \multicolumn{2}{c}{\textbf{Components}}  & \multicolumn{2}{c}{\textbf{L2P}} & \multicolumn{2}{c}{\textbf{Dual Prompt}} \\
    \cmidrule(lr){2-3} \cmidrule(lr){4-5} \cmidrule(lr){6-7}
    & $\mathcal{L}_{task}$ & $\mathcal{L}_{class}$ & Acc & Forg &
    Acc & Forg \\
    \midrule
    a) & &  & 61.57 & 9.73 & 68.13 & 4.68 \\
    b) & \checkmark & & 61.77 & 10.03 & 69.43 & 4.26 \\
    c) &  & \checkmark& 62.36 & 8.53 & 69.02 & 4.7\\
    d) & \checkmark & \checkmark & \bf{62.51} & \bf{4.2} & \bf{69.46} & \bf{4.2} \\
    \bottomrule
    \end{tabular}}
    \vspace{-3pt}
    \caption{\textbf{Ablating over \ours} on the challenging Split Imagenet-R dataset, we confirm the importance of each component of our model. We conclude that our method benefits from the introduction of language guidance at both the task level and at the class level. This performance improvement is achieved without introducing any additional learning parameters.}
    \label{tab:ablateloss}
    \vspace{-8pt}
\end{table}
}
{
\setlength{\tabcolsep}{4pt}
\renewcommand{\arraystretch}{1.2} 
\begin{table}[t]
\setlength{\aboverulesep}{0pt}\setlength{\belowrulesep}{0pt}
    \centering
    {
    \begin{tabular}{l  c c}
    \toprule
    \textbf{Text Encoder} & \textbf{Avg. Accuracy} & \textbf{Forgetting}\\
    \midrule
    RoBERTa~~\cite{liu2019roberta} & 87.04 & 5.4   \\
    BERT~\cite{Lan2019albert} & 87.11 &  \textbf{4.90}  \\
    CLIP~\cite{Radford2021clip} & \textbf{87.23} & 5.10  \\
    \bottomrule
    \end{tabular}}
    \vspace{-5pt}
    \caption{\textbf{Ablation over different text encoders.} We test our proposed method with CLIP~\cite{Radford2021clip}, BERT~\cite{devlin2018bert} and RoBERTa~\cite{liu2019roberta} text encoders. All experiments in this table are conducted on Dual Prompt~\cite{wang2022dualprompt} + LGCL. We observe that CLIP~\cite{Radford2021clip} demonstrates the highest performance. }
    \label{tab:ablate_text_encoder}
    \vspace{-10pt}
\end{table}
}

{
\setlength{\tabcolsep}{4pt}
\renewcommand{\arraystretch}{1.2} 
\begin{table}[t]
\setlength{\aboverulesep}{0pt}\setlength{\belowrulesep}{0pt}
    \centering
    {
    \begin{tabular}{l l  c c}
    \toprule
    & \textbf{Keys} & \textbf{Avg. Accuracy} & \textbf{Forgetting}\\
    \midrule
    a) & Frozen CLIP Keys & 86.15 & \textbf{3.93}  \\
    b) & Learnable Keys & \textbf{87.23} &  5.10  \\
    \bottomrule
    \end{tabular}}
    \vspace{-5pt}
    \caption{\textbf{Ablation over different keys $K_s$.} We replace the keys with CLIP~\cite{Radford2021clip} CLS tokens and use our loss function. }
    \label{tab:ablate_keys}
    \vspace{-10pt}
\end{table}
}
\subsection{Ablation on Text Transformer.}
In previous experiments, we use the text transformer from a pre-trained CLIP model. CLIP was pre-trained on images and their captions from the internet and therefore learns image-informed text embeddings. In Table~\ref{tab:ablate_text_encoder}, we additionally ablate over text transformers from pre-trained language-only models, namely BERT~\cite{Lan2019albert} and RoBERTa~\cite{liu2019roberta}. We perform this ablation with Dual Prompt + \ours on Split CIFAR-100. We observe from the table that the CLIP Text Transformer achieves the best result in Average Accuracy since it is pre-trained with both image and text data. However, we see a reasonable performance gain with Language Only pre-trained Text Encoders RoBERTa and BERT. This validates that \ours is fairly robust to the choice of text encoder. \

\subsection{Ablation on Keys of the Prompt Pool.}
We ablate over the design choice for the keys of the prompt pool in Table~\ref{tab:ablate_keys}. We ablate using the Split CIFAR-100 dataset with Dual Prompt + \ours. The keys of the prompt pool are learnable and responsible for selecting the most relevant prompt(s) for the task with a query key lookup from the $\texttt{CLS}$ feature of the Image Transformer. Therefore improving the keys can result in performance improvement. We test two different strategies here. 
In row a), we replace the keys with the CLS tokens from the CLIP Text Transformer and keep them frozen. These keys represent the targets we optimise with our $\mathcal{L}_{task}$. We observe that this, while competitive, does not reflect the performance gains of \ours over Dual Prompt. In row b) we notice that the learnable keys with our $\mathcal{L}_{task}$ achieve the best performance indicating the effectiveness of our formulation.


\section{Conclusion.}
We introduce a novel perspective of introducing language guidance in prompt-based continual learning in this work. The key intuition behind our approach is that even though the task distributions change between tasks, their label space can be mapped to the same language space. A model that can learn to map to this space can mitigate catastrophic forgetting, leading to performance improvement. We introduce language guidance at two levels; namely task-level and class-level.
At task-level, we introduce language guidance for prompt pool, where the model needs to select relevant prompts for class conditioning of a pre-trained vision transformer. By improving the key lookup of the prompt pool, we allow the model to be more robust across different tasks. To this end, we encourage the model to map the keys to its respective task-level language representation.
Secondly, we introduce language guidance at the class-level in the output feature of the vision transformer. At this stage, we incentivise the model to map the output feature to the class level language representation. Without any additional learning parameters, our method improves the performance of baseline prompt-based continual learning methods to set a new state-of-the-art.

{\small
\bibliographystyle{ieee_fullname}
\bibliography{egbib}
}

\end{document}


\title{Supplementary: Introducing Language Guidance in Prompt-based Continual Learning}

\author{Muhammad Gul Zain Ali Khan$^{1,2}$ \quad Muhammad Ferjad Naeem$^{3}$\quad Luc Van Gool$^{3}$\quad Didier Stricker$^{1,2}$\\ Federico Tombari$^{4,5}$\quad Muhammad Zeshan Afzal$^{1,2}$\\
{\small $^1$RPTU\quad $^2$DFKI\quad $^3$ETH Zürich\quad $^4$TUM\quad $^5$Google}
}

\maketitle
\ificcvfinal\thispagestyle{empty}\fi

\section{Experiment Details}
We provide experimental details of all experiments in this section. We implement \ours using PyTorch on  Ubuntu 20.0 workstation. We conduct all experiments on A100-40GB GPUs. We use a constant learning rate of 0.005 using Adam Optimizier~\cite{kingma2014adam} with $\beta_1=0.9$ and $\beta_2=0.999$. We use batch size 24 for DualPrompt~\cite{wang2022dualprompt} and batch size 16 for L2P~\cite{Wang2021l2p}. We resize all input images to 224$\times$224. \\
For L2P, we use 50 epochs for Split-ImageNet-R~\cite{wang2022dualprompt} and 5 epochs for CIFAR-100~\cite{Krizhevsky2009cifar100} datasets. For DualPrompt~\cite{wang2022dualprompt} we use 50 epochs for Split-ImageNet-R and 20 epochs for L2P~\cite{Wang2021l2p}. We set $M=10$, $L_p = 5$ and $N=5$ for L2P~\cite{Wang2021l2p} on Split-CIFAR100~\cite{Krizhevsky2009cifar100} dataset. We set $M=10$, $L_e = 20$, $L_g = 5$ and $N=1$ for DualPrompt~\cite{wang2022dualprompt} on Split-CIFAR100~\cite{Krizhevsky2009cifar100} dataset. We set $M=10$, $L_e=20$, $L_g=5$ and $N=1$ for Dualprompt~\cite{wang2022dualprompt} on Split-ImageNet-R~\cite{wang2022dualprompt} dataset. We set $M=30$,$L_p=20$ and $N=5$ for L2P~\cite{Wang2021l2p} on Split-ImageNet-R~\cite{wang2022dualprompt}. We use CLIP~\cite{Radford2021clip} text encoder ``ViT-L/14" for generating language representations ($L_t$ and $L_c$). Specifically, we use ``ViT-L/14" for L2P~\cite{Wang2021l2p} and ``ViT-L14/336px" for DualPrompt~\cite{wang2022dualprompt}. Following ~\cite{Wang2021l2p,wang2022dualprompt}, we use ViT-B/16 pre-trained backbone. \\
For Split-ImageNet-R~\cite{wang2022dualprompt}, we set 0.3 loss weight for task-level language guidance loss given in Eq. 3 and 0.7 loss weight for class-level language guidance loss given in Eq. 4. For Split-CIFAR100~\cite{Krizhevsky2009cifar100}, we set 0.4 loss weight for task-level language guidance loss given in Eq. 3 and 0.6 loss weight for class-level language guidance loss given in Eq. 4. For L2P~\cite{Wang2021l2p}, we set 0.1 loss weight for task-level language guidance loss given in Eq. 3 and 0.9 loss weight for class-level language guidance given in Eq. 4 on Split-CIFAR100~\cite{Krizhevsky2009cifar100} dataset. For Split-ImageNet-R~\cite{deng2009imagenet}, we set 0.5 loss weight for both task-level language guidance loss given in Eq. 3 and class-level language guidance loss given in Eq. 4 with L2P~\cite{Wang2021l2p}

\section{Evaluation Metrics}
We compute Average Accuracy at Task $t$ denoted by $A_t$ and Forgetting at Task $t$ denoted by $F_t$. Let $E_{t,\mathcal{T}}$ be the accuracy of task $t$ when evaluated at task $\mathcal{T}$. Then $A_t$ is shown below:
\begin{align*}
    A_t = \frac{1}{t} \sum_{\mathcal{T}}^t E_{t,\mathcal{T}}
\end{align*}
\newline
Average accuracy represents an average of the accuracy of all the tasks at a given task $t$. However, this does not represent how much the model has forgotten from the previous tasks. Forgetting $F_t$ aims to quantify the catastrophic forgetting of Neural Networks in Continual Learning. Forgetting $F_t$ formulation is given below.
\begin{align*}
    F_t= \frac{1}{t-1}\sum_{\mathcal{T}}^{t-1}max_{\mathcal{T}' \in \{1,...,t-1\}}(E_{\mathcal{T}',\mathcal{T}}-E_{t,\mathcal{T}})
\end{align*}
\section{Comparison with L2P on Prompt Length and Selection Size}
We show ablation on Prompt Length ``L" and Selection Size ``N" on L2P~\cite{Wang2021l2p} in Table. ~\ref{tab:l2p_param_ablate}. For comparison with baseline method L2P~\cite{Wang2021l2p}, we report average accuracies on Split-CIFAR100~\cite{Krizhevsky2009cifar100}. We show that \ours outperforms baseline method L2P~\cite{Wang2021l2p} on most configurations. \ours achieves the highest accuracy at N=5 and L=20. However, the highest difference in performance is with N=20 and L=20 where \ours outperforms baseline method L2P~\cite{Wang2021l2p} by 1.51\%. \ours is comparable to baseline method L2P~\cite{Wang2021l2p} where L2P~\cite{Wang2021l2p} outperforms \ours with highest difference of 0.8\% at N=1 and L=10. We observe that \ours is fairly robust to changes in prompt configuration since \ours outperforms baseline method L2P~\cite{Wang2021l2p} on most configurations and produces comparable results in configurations where L2P~\cite{Wang2021l2p} outperforms \ours. We achieve this consistent performance without any addition in the number of parameters. 
\begin{table}[h]
\centering
\resizebox{0.7\columnwidth}{!}{
\begin{tabular}{|l|l|l|l|l|}
\hline
\diagbox[]{L}{N} & 1  & 5  & 10 & 20 \\ \hline
1                  & \textbf{77.63}/77.01 & 81.49/\textbf{82.49} & 82.92/\textbf{83.39} & \textbf{83.34}/83.15 \\ \hline
5                  & 82.24/\textbf{82.88} & \textbf{83.85}/83.50 & \textbf{83.90}/83.39 & \textbf{83.79}/82.84         \\ \hline
10                 & 82.48/\textbf{83.28} & \textbf{83.68}/83.18 & \textbf{83.52}/83.13 & \textbf{81.98}/81.84         \\ \hline
20                 & \textbf{83.86}/83.40 & \textbf{84.01}/82.60 & \textbf{82.65}/81.10 & \textbf{81.19}/79.65         \\ \hline
\end{tabular}
}
\caption{\myparagraph{Ablation on Prompt Length ``L" and Selection Size ``N" on L2P~\cite{Wang2021l2p}.} We compare \ours with baseline method L2P~\cite{Wang2021l2p} on ``L" and ``N". We report average accuracy on Split-CIFAR100~\cite{Krizhevsky2009cifar100}. We report the average accuracy of L2P~\cite{Wang2021l2p} and \ours both. The first result in each cell is \ours average accuracy followed by a / and then L2P~\cite{Wang2021l2p} average accuracy. Results for L2P~\cite{Wang2021l2p} are taken from \cite{Wang2021l2p}. Higher average accuracy is in \textbf{bold}. We keep prompt pool size ``M" constant at 20 and all other hyperparameters. We show that \ours consistently outperforms baseline method L2P~\cite{Wang2021l2p} on most configurations.}
\label{tab:l2p_param_ablate}
\end{table}


\section{Limitations}
\ours does not explore other modalities since \ours builds on previous prompting based continual learning methods ~\cite{Wang2021l2p,wang2022dualprompt}. \ours assumes that robust VIT~\cite{Dosovitskiy2020VIT} based pre-trained feature extractor is available. \ours also assumes that a pre-trained text encoder that can generate robust text embeddings of the classes is available during training. Furthermore, \ours explores language modelling on VIT~\cite{Dosovitskiy2020VIT} based networks. We leave the exploration of \ours and language modelling on conventional ConvNet-based networks as future work. While we have shown that Incremental Class setting of continual learning can benefit from language modelling and \ours, we leave the exploration of \ours in other domains of continual learning, such as task-agnostic continual learning, as future work. 

{\small
\bibliographystyle{ieee_fullname}
\bibliography{egbib}
}